\documentclass[conference]{lib/IEEEtran}
\IEEEoverridecommandlockouts
\usepackage{cite}
\usepackage{amsmath,amssymb,amsfonts}
\usepackage{algorithmic}
\usepackage{graphicx}
\usepackage{textcomp}
\usepackage{xcolor}
\def\BibTeX{{\rm B\kern-.05em{\sc i\kern-.025em b}\kern-.08em
    T\kern-.1667em\lower.7ex\hbox{E}\kern-.125emX}}
\usepackage{enumitem,bigstrut,multirow}

\usepackage{bm}
\usepackage{float}
\usepackage{booktabs}
\usepackage{array}
\usepackage{tikz}
\usepackage{graphicx}
\usepackage{subfig}
\usepackage{amsmath,amsfonts,amssymb,amsthm}

\usepackage{wrapfig} % For inline figures

\newcommand{\ie}{\textit{i.e.}}
\newcommand{\eg}{\textit{e.g.}}

\begin{document}

\title{
Towards Reproducible Learning-based Compression
}

\author{\IEEEauthorblockN{Jiahao~Pang, Muhammad~Asad~Lodhi, Junghyun~Ahn, Yuning~Huang, Dong~Tian}
\IEEEauthorblockA{\textit{InterDigital}, New York, USA \\
\{jiahao.pang, muhammad.lodhi, junghyun.ahn, yuning.huang, dong.tian\}@interdigital.com}}

\maketitle

\begin{abstract}
A deep learning system typically suffers from a lack of reproducibility that is partially rooted in hardware or software implementation details. 
The irreproducibility leads to skepticism in deep learning technologies and it can hinder them from being deployed in many applications. 
In this work, the irreproducibility issue is analyzed where deep learning is employed in compression systems while the encoding and decoding may be run on devices from different manufacturers. 
The decoding process can even crash due to a single bit difference, \eg, in a learning-based entropy coder. 
For a given deep learning-based module with limited resources for protection, we first suggest that reproducibility can only be assured when the mismatches are bounded.
Then a safeguarding mechanism is proposed to tackle the challenges.
The proposed method may be applied for different levels of protection either at the reconstruction level or at a selected decoding level.
Furthermore, the overhead introduced for the protection can be scaled down accordingly when the error bound is being suppressed. 
Experiments demonstrate the effectiveness of the proposed approach for learning-based compression systems, \eg, in image compression and point cloud compression.
\end{abstract}


% -------------------------------------------------------------------------
\section{Introduction}
\label{sec:intro}
For deep learning systems, there exists a well-known problem which is reproducibility, especially \emph{inference reproducibility}.
That is, the same trained neural network model may produce results with minor differences on different hardware platforms such as different Graphics Processing Units (GPUs), or different software platforms such as PyTorch and TensorFlow.
A major reason for irreproducibility is the non-determinism in the platforms when executing operations in parallel, which is sensitive to computation orders. 
Another reason causing such mismatches is the different implementations across different software packages of elementary operators used to build neural network models, \eg, the fully-connected layers and the convolutional layers.
\begin{figure}[t]
    \centerline{\includegraphics[width=1\linewidth]{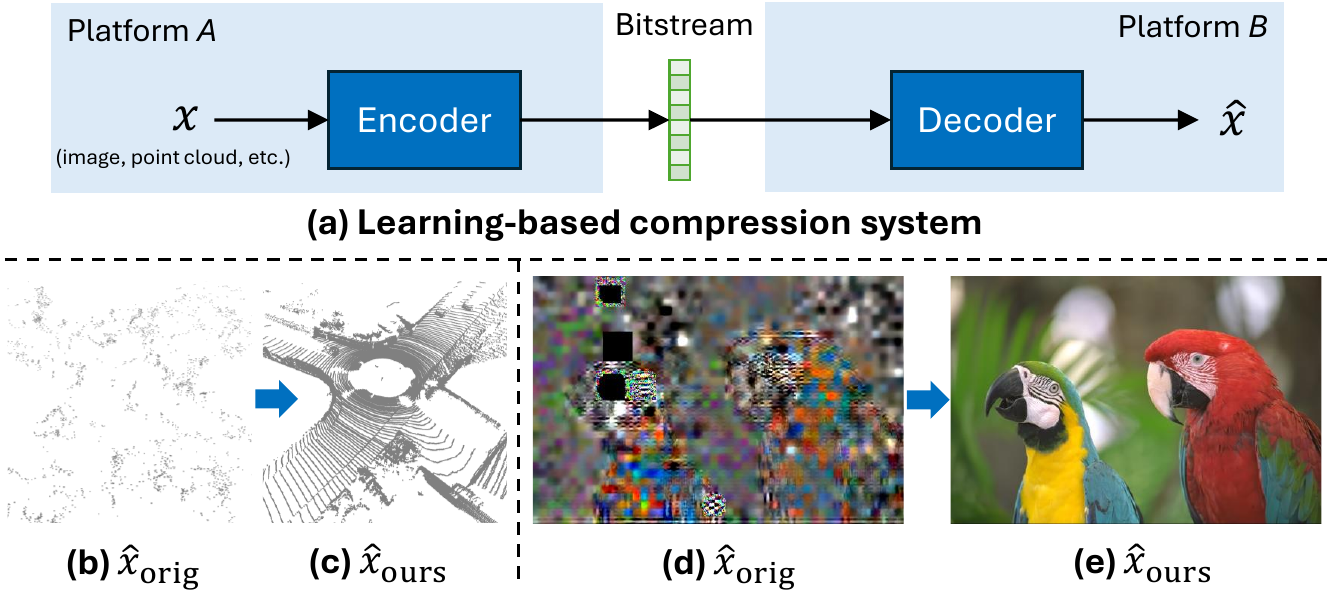}}
    \caption{(a) We ensure reproducibility for learning-based compression systems to achieve interoperability across platforms. (b) Without protection, a lossless point cloud decoder on an H100 GPU fails to decode a point cloud encoded with an A100 GPU. (c) With our proposal, the decoder decodes correctly. In (d)(e), a similar example of lossy image coding is provided. In both cases, the bitstream overhead is around $1\%$.}
    \label{fig:demo}
\end{figure}

Such irreproducibility might be tolerable for some applications. 
However, learning-based compression systems become vulnerable if the outcomes are not reproducible---there is no guarantee of interoperability between encoders and decoders on different devices/platforms.
For example, when neural network models are used for entropy coding, \eg, to predict probabilities or hyper-prior parameters for encoding/decoding of data, a minor difference between encoding and decoding can not only lead to significantly different data being reconstructed but may even crash the decoding (see the examples in Fig.~\ref{fig:demo}).

There exist some techniques to alleviate the irreproducibility of neural network models.
A straightforward approach is network quantization which replaces the vanilla 32-bit floating point data types with lower-precision data types, \eg, 16-bit floating point. 
Network quantization is mainly introduced to reduce the computational and memory costs of inference~\cite{nagel2021white}.
Fortunately, it also brings more consistent behavior across platforms because slightly different values may become the same after quantization.
In addition, certain implementation tricks may also mitigate the irreproducibility. 
For instance, the popular cuDNN library~\cite{chetlur2014cudnn} determines the fastest convolution algorithm on the fly.
A different algorithm thus may be selected on a different platform, leading to slightly different results.
To ensure the same algorithm is always selected, one may disable the search of the fastest algorithm~\cite{pytorch2023reproducibility}.
However, unless the hardware and software platforms are strictly aligned, all the mentioned techniques cannot guarantee inference reproducibility.
In \cite{balle2018integer}, Balle~\textit{et al.} particularly introduce an approach to address the reproducibility issue, but under the limited context of a latent-variable model for quantized (integer) networks.

In contrast, to the best of our knowledge, our work is a \emph{first} attempt in the literature to ensure inference reproducibility for learning-based compression systems.
Particularly, starting from an error-bounded assumption across platforms, we quantize the output of the critical modules of the learning-based encoder.
Then a safeguarding bitstream is encoded to guarantee the same results can be reproduced by the learning-based decoder on a target platform so that the decoding process runs correctly.

Our proposal sheds light on interoperability across heterogeneous platforms.
It can be applied as a \emph{plug-and-play} add-on for existing learning-based compression systems---no updates or finetuning to a pretrained neural network model is required. 
Moreover, as to be demonstrated, it is a \emph{generic} approach applicable for lossless or lossy coding, and for different modalities such as image/video and point cloud.

%------------------------------------------------------------------------
\section{Reproducible Learning-based Compression}
\label{sec:reproducible_learning}
This section presents our methodology to achieve reproducible learning-based compression.
It first dives into the reproducibility issue in compression, followed by our proposal to safeguard reproducibility.

\subsection{Reproducibility in Compression}

\begin{figure}[t]
    \centerline{\includegraphics[width=0.7\linewidth]{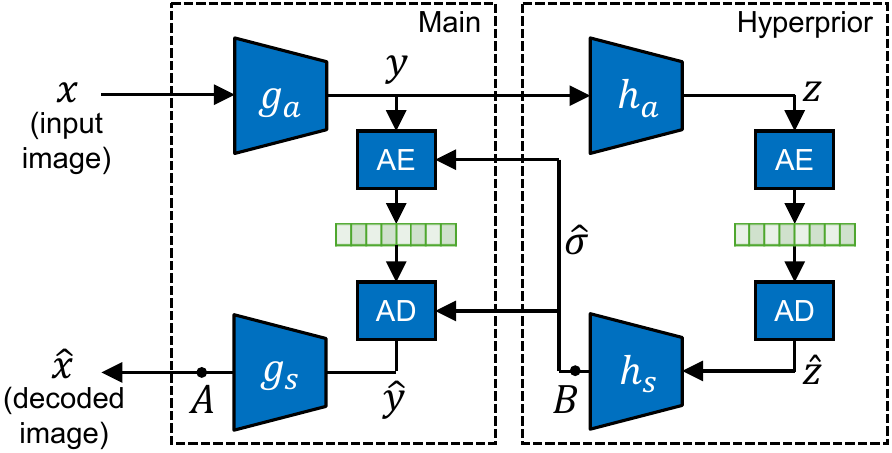}}
    \caption{Reproducibility in a hyperprior-based image codec. Point $A$ is to check \emph{reconstruction reproducibility}. Point $B$ is to check \emph{decoding reproducibility.}
    }
    \label{fig:hyperprior_reproducibility}
\end{figure}

To study reproducibility in learning-based compression, we take lossy image compression with scale hyper-prior as an example codec \cite{balle2018variational}, where at the main branch, image features are extracted and arithmetically encoded, then the decoder reconstructs the image based on the arithmetically decoded features (Fig.~\ref{fig:hyperprior_reproducibility}).

For what we called \emph{reconstruction reproducibility}, it concerns the final reconstruction, \ie, the output image $\hat{x}$ (at point~$A$) is required to be bit-wise matched across platforms. 
Note that any randomness in the feature extraction $g_a$ does not undermine the reproducibility but only coding performance.
Instead, a vulnerable reconstruction procedure $g_s$ contaminates the decoded results and needs to be protected for reconstruction reproducibility.

The image codec in with scale hyperprior Fig.~\ref{fig:hyperprior_reproducibility} is further composed by a hyperprior branch, where hyperprior features $z$ are extracted and encoded/decoded.
The hyperprior feature is used to synthesize hyperprior parameters (\eg, Gaussian parameters such as variances $\hat{\sigma}^2$) which assist the arithmetic coding of image features $y$.
Thus, it is unfair to just blame $g_s$ alone for reconstruction reproducibility.

Similar to the earlier case, the hyper-analysis block $h_a$ won't be a problem for reproducibility, but the hyper-synthesis block $h_s$ is.  
If there is any mismatch, even a single bit difference, happens in $\hat{\sigma}$ (the output of hyper-synthesis), the arithmetic decoding of the image features may crash, and hence the whole decoder (See Fig.~\ref{fig:demo}d). 
Consequently, we introduce \emph{decoding reproducibility} for a learning-based arithmetic coder, \eg, defined at point~$B$ in Fig.~\ref{fig:hyperprior_reproducibility}, while not limited just to hyperprior models \cite{balle2018variational}.
Apparently, the decoding reproducibility \emph{shall be assured} to avoid fatal decoding problems by protecting the hyper-synthesis model.
For reconstruction reproducibility, the reconstruction process needs to be additionally protected. 

\begin{figure}[t]
    \centerline{\includegraphics[width=0.7\linewidth]{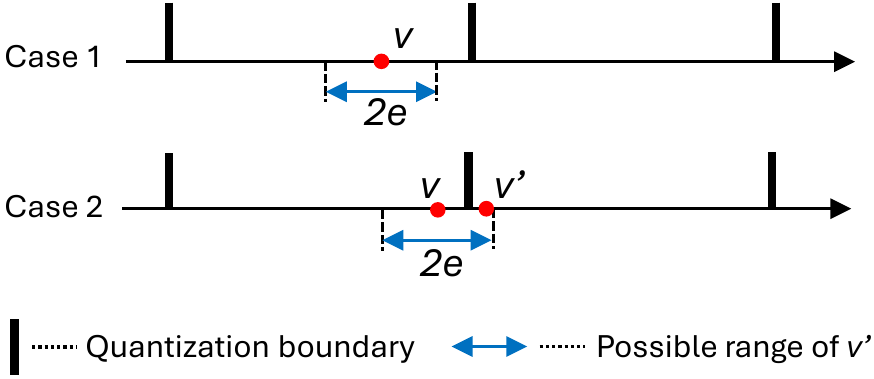}}
    \caption{Quantization of an error-bounded variable.}
    \label{fig:quantize}
\end{figure}

Suppose in a learning-based compression system, a module $\mathcal{M}$ is deployed to produce a scalar variable $v$ during the encoding/decoding of a signal (image, video, point cloud, etc.).
When $v$ dictates part of the entropy coding, \ie, it directly affects how the signal is being decoded, this scalar variable as well as the module $\mathcal{M}$ is identified as \emph{critical} for decoding reproducibility.
For example, in Fig.~\ref{fig:hyperprior_reproducibility}, the variances are critical variables and $h_s$ is a critical module.
In the example of tree-based lossless point cloud compression (PCC), the occupancy probabilities of the voxels to be coded are critical variables, and the estimation network predicting probabilities is a critical module~\cite{wang2022sparse,fu2022octattention}.

The objective of our work is to address the decoding reproducibility issue by protecting the critical modules $\mathcal{M}$ so that the critical variables generated by them can be reproduced.
Theoretically, our method can also be used for reconstruction reproducibility by paying a higher overhead in bitrate. 
\begin{figure*}[t]
  \centering\scriptsize
  \captionsetup[subfigure]{width=230pt}
  \subfloat[Encoder]{\includegraphics[width=230pt]{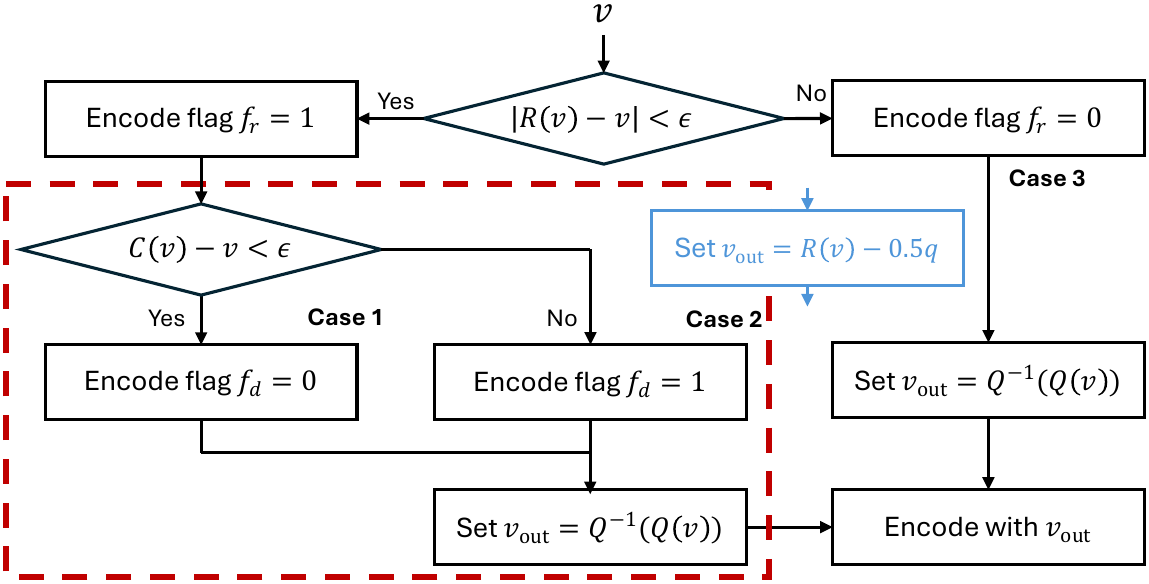}\label{fig:basic_enc}}\hspace{20pt}
  \subfloat[Decoder]{\includegraphics[width=230pt]{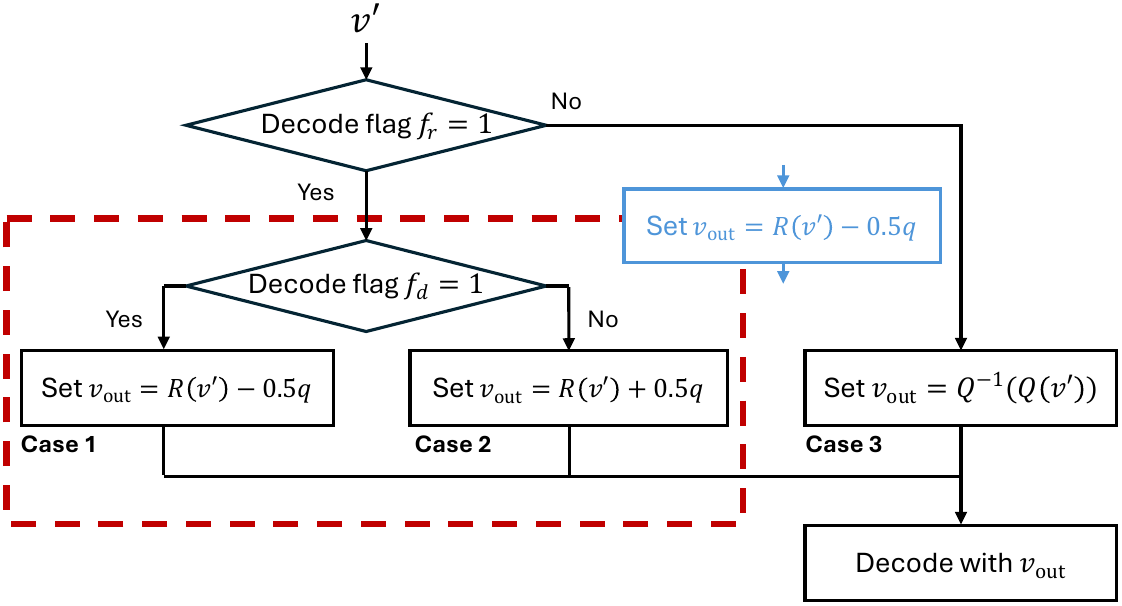}\label{fig:basic_dec}}
  \caption{Encoder and decoder flowcharts of our proposal. The steps in black---utilize the direction flag $f_d$ to reproduce $Q^{-1}(Q(v))$ on the decoder. Replacing the circled parts by the associated steps in blue color---the simplified proposal without flag $f_d$.}
  \label{fig:basic}
\end{figure*}
\vspace{-5pt}

\subsection{Quantization for Error-bounded Platforms}\label{ssec:repro_compression}
We first introduce our starting point, the error-bounded assumption.
Suppose on the encoder platform, critical module $\mathcal{M}$ generates a critical scalar variable $v$ that needs to be reproduced. 
Due to potential mismatches between the encoder and decoder platforms, a different scalar, $v'$, is generated by the decoder.
We assume that given module $\mathcal{M}$ and a pair of known encoder and decoder platforms, the error between $v$ and $v'$ is \emph{bounded}, that is, $ \left| v - v' \right| \le e$ where $e \ge 0$ is the maximum achievable error.

Given a bounded error between $v$ and $v'$, a first attempt to reduce their difference is to apply quantization as discussed in Section~\ref{sec:intro}.
Without loss of generality, suppose uniform quantization with a step size $q$ is applied, \ie,
\begin{equation}\label{eq:quantization}
    Q(v) = \left \lfloor \frac{v}{q} + s \right \rfloor,
\end{equation}
where $\left \lfloor \cdot \right \rfloor$ is the flooring operator, $s\in[0, 1)$ is an offset, and we let $q > 4e$.

Fig.~\ref{fig:quantize} demonstrates the quantization of $v$ and $v'$.
As shown in Case~1, if $v$ has a distance of at least $e$ away from the quantization boundaries, applying quantization alone is sufficient to make $v$ and $v'$ the same because both of them fall in the same quantization bin. 
In contrast, as shown in Case~2 of Fig.~\ref{fig:quantize}, when $v$ falls within the vicinity of the quantization boundaries (within a distance $e$), applying quantization has no guarantee to make $v'$ the same as $v$ because $v'$ may fall into a different bin from $v$.
This holds true no matter how large the quantization step size becomes.

However, we have two key observations via this analysis.
i)~By quantizing the critical variable $v$, $Q(v)$ and $Q(v')$ becomes at most $1$ apart, \ie, $ \left| Q(v) - Q(v') \right| \le 1$, because the quantization step size (length of the quantization bin) $q > 4e$.
ii)~If $q$ gets larger, it is more likely that $v$ is not in the vicinity of the quantization boundaries.
Note that these two observations hold even when non-uniform quantization is used.
They motivate the next step of our proposal.

\subsection{Proposed Safeguarding Approach}\label{ssec:safeguarding}
In a nutshell, to ensure the decoder reproduces a $Q(v')$ the same as $Q(v)$, we let the encoder signal an extra boolean flag indicating whether $v$ is risky, \ie, whether it falls within a distance $e$ from its closest the quantization boundary (\eg, Case~2 of Fig.~\ref{fig:quantize}).
Additionally, for a critical variable $v$ that is risky, the encoder signals another boolean flag indicating $v$ should be on the left or the right side of its closest quantization boundary.
According to these extra signalings, the decoder can recover $Q(v)$ \emph{exactly} from $v'$.

Instead of directly working with the maximum achievable error $e$, we pick a \emph{maximum tolerable error} $\epsilon$ to determine whether a critical variable is risky, where $q>4\epsilon$.
Therefore, when the actual maximum achievable error $e$ is less than the maximum tolerable error $\epsilon$, the bitstream generated by our encoder is guaranteed to be correctly decoded by the decoder.
In other words, the bitstream is \emph{reproducibility compatible} with any decoder platform with a maximum achievable error $e<\epsilon$ respective to the encoder platform.

In Fig.~\ref{fig:basic}, the blocks in black color show the encoder and decoder flowcharts of our proposal.
Note that $C(v)$ is the ceiling operator which computes the upper quantization boundary associated with $v$. 
Likewise, we define the flooring operator $F(v)$ which computes the lower quantization boundary.
$R(v)$ is the rounding operator which rounds $v$ to the closest quantization boundary, \ie,
\begin{equation}
    R(v)=\left\{\begin{matrix}
C(v), & v-F(v)>C(v)-v, \\ 
F(v), & \textrm{otherwise}.
\end{matrix}\right.
\end{equation}
$Q^{-1}(\cdot)$ is dequantization.
It outputs a value at the center of the associated quantization bin.
For uniform quantization,
\begin{equation}
Q^{-1}(n)=(n+0.5-s)q.
\end{equation}

\begin{figure}[t]
    \centerline{\includegraphics[width=0.7\linewidth]{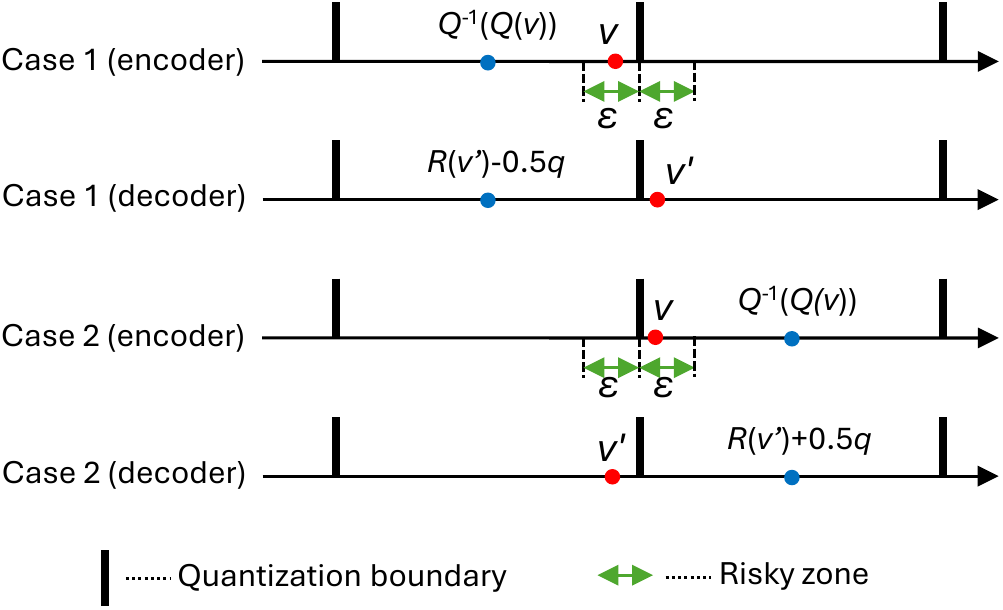}}
    \caption{Error correction when a variable falls in the risky zone.}
    \label{fig:alg_basic}
\end{figure}

\textbf{Encode}: As shown in Fig.~\ref{fig:basic_enc}, If $v$ is close to its rounded value $R(v)$ within a distance $\epsilon$ (in the ``risky'' zone), a risky flag $f_r$ associated with $v$ will be set to $1$ and encoded into the bitstream---indicating that $v$ needs extra care to ensure reproducibility on the decoder side.
Next, if $C(v)-v<\epsilon$, \ie, $v$ is closer to $C(v)$ rather than $F(v)$, an additional flag $f_d$ about direction is set to be $0$ and also encoded in the bitstream.
It indicates the dequantized version of $Q(v)$ should be on the left side of its rounded value, \ie, $Q^{-1}(Q(v))=R(v)-0.5q$ (see Case~1 (encoder) in Fig.~\ref{fig:alg_basic}).
Otherwise, it implies $v$ is closer to its floored value $F(v)$, and a direction flag $f_d=1$ will be encoded into the bitstream.
It indicates the dequantized version of $Q(v)$ should be on the right side of its rounded value, \ie, $Q^{-1}(Q(v))=R(v)+0.5q$ (See Case~2 (encoder) in Fig.~\ref{fig:alg_basic}).
If $v$ is at least having a distance $\epsilon$ apart from $R(v)$, quantizing $v$ alone already guarantees $Q(v)$ and $Q^{-1}(Q(v))$ can be reproduced on the decoder, so no extra action is required for protection.

In the end, instead of using $v$ produced by $\mathcal{M}$ to continue the encoding, we let $v_{\rm out}=Q^{-1}(Q(v))$, and use $v_{\rm out}$ to perform all subsequent encoding steps.
Note that this proposed procedure needs to be performed for all critical values generated by $\mathcal{M}$ so as to protect the reproducibility of $\mathcal{M}$.

\textbf{Decode}: The decoding associated with the encoding procedure is shown in Fig.~\ref{fig:basic_dec}.
On the decoder platform, the value $v'$ generated by $\mathcal{M}$ may have a minor difference compared to $v$ produced on the encoder, and the objective is to reproduce the value $Q^{-1}(Q(v))$ to have a correct decoding process.
The decoder first checks if this critical variable is \emph{risky} by checking its associated risky flag $f_r$.
If not ($f_r=0$), the variable $v$’s value (on encoder) is already in a safe range and $Q^{-1}(Q(v'))=Q^{-1}(Q(v))$ is guaranteed, so we compute $v_{\rm out}=Q^{-1}(Q(v'))$ directly.

If $f_r=1$, it indicates a mismatch may happen between $Q(v)$ and $Q(v')$, especially when $v'$ and $v$ fall at different sides of a quantization boundary, we have $\left| Q(v) - Q(v') \right| = 1$. 
However, because $4\epsilon < q$, even in the worst case, $R(v)=R(v')$ still holds.
Thus, the direction flag $f_d$ is used to reproduce $Q^{-1}(Q(v))$.
If $f_d=0$ is decoded, implying that on the encoder, the value $v_{\rm out}$ falls on the left side of $R(v)$, so the decoder needs to output a $v_{\rm out}$ also on the left side of $R(v')$, given by $R(v')-0.5q$ (Case~1 (decoder) of Fig.~\ref{fig:alg_basic}).
If $f_d=1$ is decoded, meaning that on the encoder, the value $v_{\rm out}$ falls on the right side of $R(v')$.
So the decoder outputs $v_{\rm out}= R(v') + 0.5q$ (Case~2 (decoder) of Fig.~\ref{fig:alg_basic}). 

In the end, instead of using the value $v'$ produced by $\mathcal{M}$ on the decoder to continue the decoding, we use $v_{\rm out}$ to accomplish all the subsequent decoding steps.

\begin{figure}[t]
    \centerline{\includegraphics[width=0.8\linewidth]{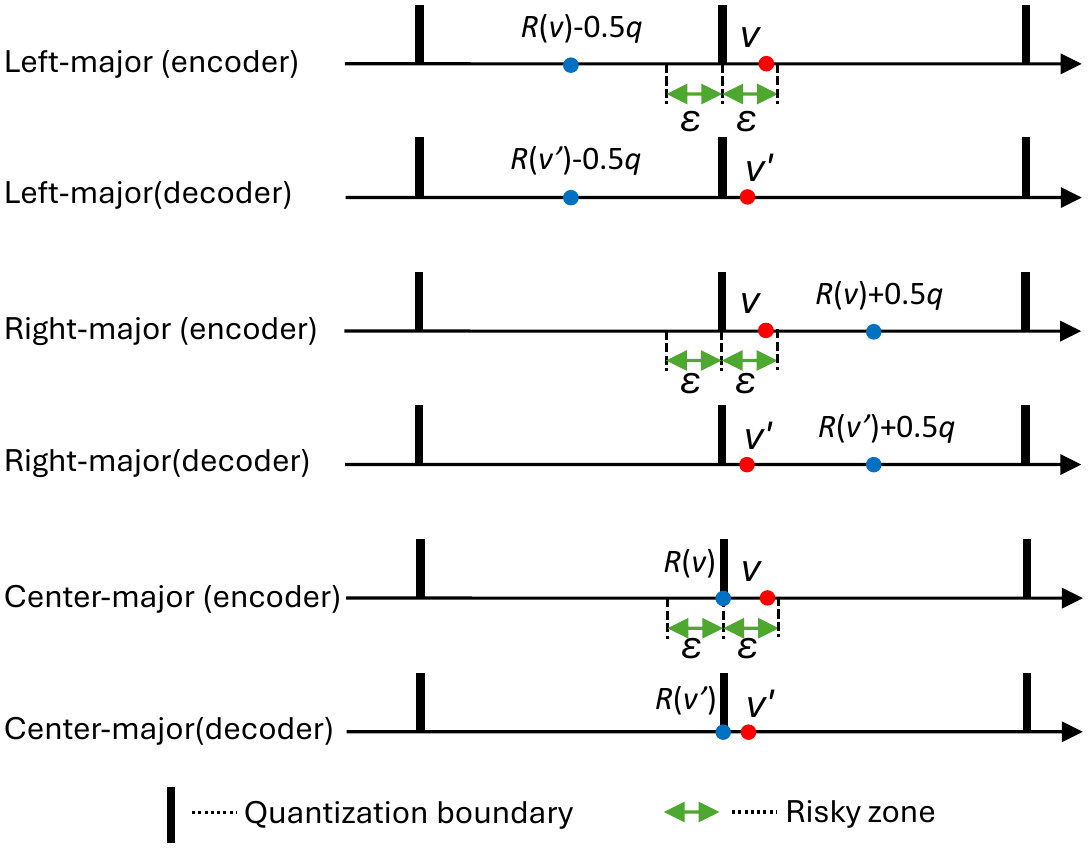}}
    \caption{Error correction without $f_d$ when $v$ is in the risky zone.}
    \label{fig:alg_advance}
\end{figure}

\subsection{Simplification}\label{ssec:reproducible_learning:refined}
We additionally simplify our proposal by skipping the coding of the direction flag $f_d$.
This is possible because when $f_r=1$, the $R(v)$ on the encoder and $R(v')$ on the decoder have the same value.
In Fig.~\ref{fig:basic_enc}, we simplify the encoder by replacing the circled steps with the step highlighted in blue color, the decoder simplification is similarly done in Fig.~\ref{fig:basic_dec}.
On the encoder side, if $\left| R(v) - v \right| < \epsilon$, \ie, $v$ falls into the risky zone, we encode its risky flag $f_r=1$, and set the output as $v_{\rm out}=R(v)-0.5q$ (the step in blue), which is the bin center to the left of its rounded value.
We call this a \emph{left-major} approach, as illustrated in Fig.~\ref{fig:alg_advance} (Left major (encoder)).
This left-major approach is less accurate because when $v$ is located on the right side of the rounded value, it is shifted to a bin center that is not closest to $v$. 
However, it avoids sending the direction flag $f_d$.
On the decoder side (Fig.~\ref{fig:basic_dec}), if a critical variable $v'$ is not risky ($f_r=0$), the output is $v_{\rm out}=Q^{-1}(Q(v'))$. 
Otherwise, we set the output as $v_{\rm out}=R(v')-0.5q$ (the step in blue) to align with the encoder.

A similar \emph{right-major} approach can also be performed wherein the encoder output is always set to be $v_{\rm out}=R(v)+0.5q$ for risky critical variables.
To have higher precision of $v_{\rm out}$ that is closer to the input $v$, a \emph{center-major} approach can also be adopted.
In this case, the encoder output is simply $v_{\rm out}=R(v)$.
However, the $v_{\rm out}$ produced by this scheme does not fall at the bin centers but at the quantization boundaries.
Illustrative examples are also provided in Fig.~\ref{fig:alg_advance}.

We encode all the binary risky flags $f_r$ as an extra bitstream called the \emph{safeguarding bitstream}.
With all the binary risky flags, we compute the probability of them being $0$ (denoted by $p_0$), then arithmetically encode the risky flags with $p_0$, leading to the safeguarding bitstream.
Note that $p_0$ itself also needs to be encoded so that the safeguarding bitstream can be decoded by the decoder.
Empirically, we find that it is sufficient to let $\epsilon$ be much smaller than $10^{-5}$ (see Section~\ref{sec:experiments}).
Additionally, a quantization step $q$ at the magnitude of $10^{-3}$ can be used so that $q\gg 4\epsilon$. 
Then according to observation ii) in Section~\ref{ssec:repro_compression},
most of the risky flags should be $0$'s because very likely, the critical variables fall out of the risky zones (Fig.~\ref{fig:alg_basic} and Fig.~\ref{fig:alg_advance}).
Therefore, $p_0$ should be close to $1$, leading to a relatively \emph{small} safeguarding bitstream.

%------------------------------------------------------------------------
\section{Example Applications}
\label{sec:applications}
This section presents more details when our approach is used in two representative learning-based compression models: scale hyperprior for lossy image/video compression, and deep octree coding for lossless point cloud compression.

\subsection{Hyperprior Model for Image Compression}\label{ssec:hyperprior}

When the proposed safeguarding approach in Section~\ref{ssec:safeguarding} or in Section~\ref{ssec:reproducible_learning:refined} is applied to ensure the \emph{decoding reproducibility} on the hyperprior model (point $B$ in Fig.~\ref{fig:hyperprior_reproducibility}), two additional designs are described.

Since the variance $\hat{\sigma}^2$ to be protected cannot be negative, negative numbers generated in the computation are always clipped to $0$.
Thus, any values falling in $[0,\epsilon]$ do not require protection because they will not be confused with negative values.

Moreover, the variance $\hat{\sigma}^2$ in the hyperprior model would be quantized during arithmetic coding. 
Rather than uniform quantization using a constant quantization step $q$ in Section~\ref{sec:reproducible_learning}, a non-uniform quantization is performed, where the quantization boundaries are learned during training.
Therefore, we directly use the learned quantization boundaries to determine whether a $\hat{\sigma}^2$ is risky, and signal the risky flag $f_r$ accordingly.
For a hyperprior model which also learns/outputs the mean value, protection on the mean is also necessary.

\subsection{Deep Octree Model for Point Cloud Compression}\label{ssec:doc}
In this subsection, we show that the \emph{decoding reproducibility} is not limited to hyperprior models in compression.
A voxelized point cloud with $n$-bit precision can be represented via an octree decomposition structure. 
Starting from a root node representing the whole 3D space of size $2^n\times 2^n\times 2^n$, each node is equally split along all the $x$-, $y$-, and $z$- directions, leading to $2^3=8$ voxels.
For any voxel containing at least a point, it is marked to be occupied (represented by $1$); otherwise, it is marked to be empty (represented by $0$). 
By iterating this step, multiple levels of details (LoD) of the input point cloud are represented. 
The split of occupied voxels at each LoD is repeated until the last one ($n$-th LoD), then the leaves of the octree represent the point cloud. 
Thus, lossless coding of a point cloud can be achieved via coding the octree.

\begin{figure}[t]
    \centerline{\includegraphics[width=1\linewidth]{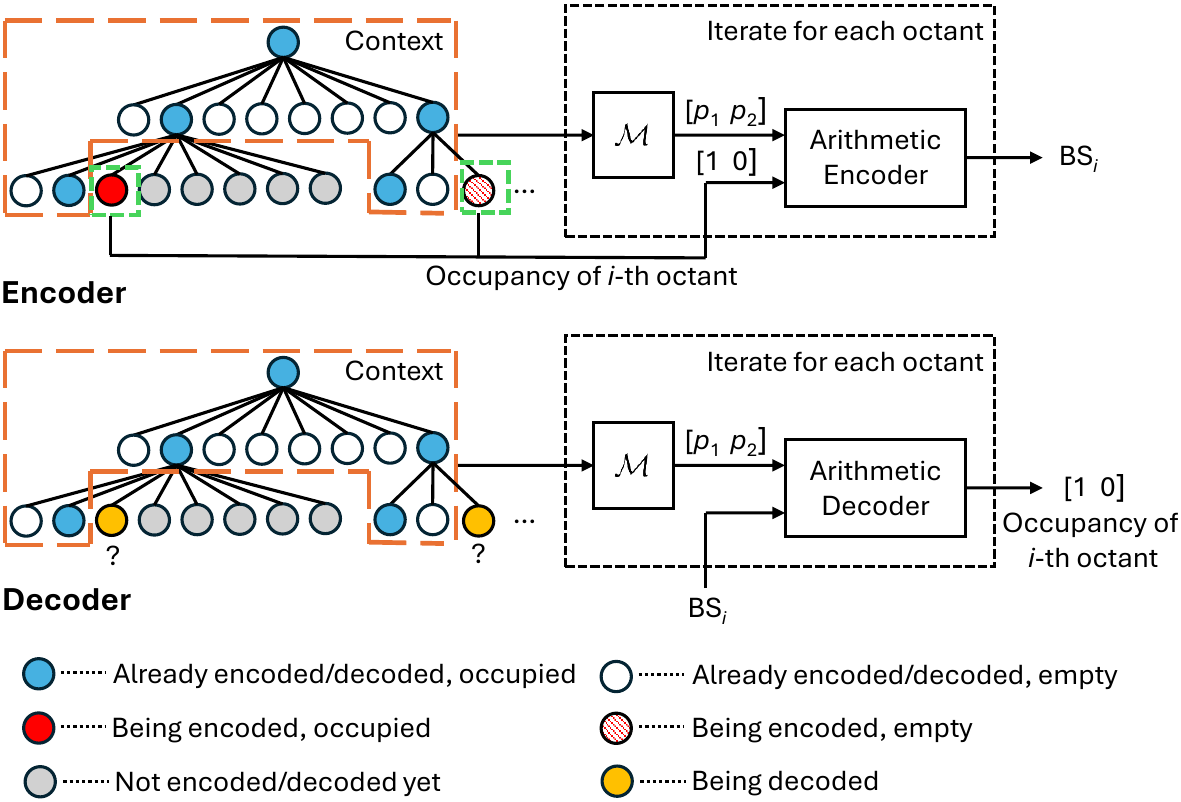}}
    \caption{A deep octree coding model. The module $\mathcal{M}$ predicting occupancy probabilities is critical and needs to be protected.}
    \label{fig:doc}
    % \vspace{-8pt}
\end{figure}

The coding of octree starts from the root voxel. We encode a sequence of occupancy bits representing the voxels at a current LoD, then we proceed to encode the next LoD.
As shown in Fig.~\ref{fig:doc}, The encoding/decoding of one octree level is split into $8$ iterations where each step encodes/decodes voxels at the $i$-th octant with respective to their parents.
To do so, a neural network model $\mathcal{M}$ takes the already encoded/decoded ancestral and sibling nodes (context) as input, then predicts the occupancy probabilities $\left\{ p_j\right\}_{j=1}^m$ of the child voxels (suppose there are $m$ child voxels) at the $i$-th octant.
The probabilities are then used to guide the arithmetic encoding/decoding.

Suppose there is a mismatch in the probabilities computed between the encoder and decoder, it will lead to wrongly decoded voxel occupancies.
Then such an error propagates to all subsequent LoDs, leading to poor decoded quality (Fig.~\ref{fig:demo}b) or even crashing the decoding process.
To protect the learning-based probability prediction $\mathcal{M}$, we propose to apply the approaches presented in Section~\ref{sec:reproducible_learning} over the estimated probability in each octant encoding/decoding iteration.

To ease the coding process, the simplified approach in Section~\ref{ssec:reproducible_learning:refined} is applied.
Particularly, the \emph{center-major} encoder is applied to digest each of the $\left\{ p_j\right\}_{j=1}^m$, and generates a safeguarding bitstream for those probability values that are risky.
On the decoder side, the risky flags are decoded along with processing the probabilities $\left\{ p_j\right\}_{j=1}^m$ with our approach presented in Section~\ref{ssec:reproducible_learning:refined}.
Note that the decoder is also configured to be \emph{center-major} decoder so as to align with the behavior of the encoder.

For simplicity, we let the offset $s$ in Eq.~\eqref{eq:quantization} be $0$, and also pick $q$ in the form of $1/k$ with $k$ being an integer so that both $0$ and $1$ are quantization boundaries.
Furthermore, the probability values fall in the range $[0,1]$. Hence, the probability values are clipped first. That is, no protection overhead is required for values in $[0,\epsilon]$ and $[1 - \epsilon, 1]$.

%------------------------------------------------------------------------
\section{Experimentation}
\label{sec:experiments}
\begin{table}[t]
  \centering\scriptsize
  \caption{Percentage of increment ($\%$) with different maximum tolerable errors $\epsilon$ for deep octree coding. Original bit-per-point (bpp) values are provided under the point cloud types.}
    \begin{tabular}{c|c||ccccc}
    \hline
    \multirow{2}[4]{*}{Point cloud} & \multicolumn{1}{c||}{\multirow{2}[4]{*}{Step size $q$}} & \multicolumn{5}{c}{Maximum tolerable error $\epsilon$} \bigstrut\\
\cline{3-7}    \multicolumn{1}{c|}{} &   & 1e-5 & 5e-6 & 1e-6 & 5e-7 & 1e-7 \bigstrut\\
    \hline
    \hline
    Dense, 10 bit & 0.004 & 14.57 & 8.46 & 2.82 & 2.02 & 1.26 \bigstrut[t]\\
     (0.68 bpp) & 0.008 & 8.54 & 5.33 & 2.41 & 2.00 & 1.63 \\
    \hline
    Dense, 11 bit & 0.004 & 13.77 & 7.82 & 2.37 & 1.59 & 0.86 \bigstrut[t]\\
     (0.62 bpp) & 0.008 & 7.88 & 4.82 & 2.04 & 1.65 & 1.31 \\
    \hline
    Dense, 12 bit & 0.004 & 11.71 & 6.57 & 1.82 & 1.13 & 0.49 \bigstrut[t]\\
     (0.93 bpp) & 0.008 & 6.69 & 3.92 & 1.40 & 1.05 & 0.72 \\
    \hline
    Sparse, 18 bit & 0.004 & 9.25 & 5.35 & 1.74 & 1.20 & 0.73 \bigstrut[t]\\
     (16.86 bpp) & 0.008 & 5.56 & 3.45 & 1.54 & 1.26 & 1.01 \\
    \hline
    \end{tabular}%
  \label{tab:doc}%
\end{table}%
% (0.72 bpp) (0.64 bpp) (0.97 bpp) (17.39 bpp)

We apply our simplified proposal in Section~\ref{ssec:reproducible_learning:refined} to the two applications in Section~\ref{sec:applications}.
Without safeguarding reproducibility, their decoders most likely fail to produce reasonable results when operating on a platform different from the encoder (see Fig.~\ref{fig:demo}).
For instance, we observe that a tiny mismatch in the (quantized) variance in the hyperprior model may lead to a reconstruction with only $10$~dB.
However, usage of our proposal brings additional costs, \ie, the safeguarding bitstream and the necessity to quantize the values to be protected.
Thus, we focus on investigating these costs, followed by more discussions.

\subsection{Overhead Analysis}
\textbf{Hyperprior for image compression}:
We first demonstrate our proposal for lossy image compression with a scale hyperprior model~\cite{balle2018variational}.
The neural network training is performed with randomly extracted and cropped patches from the Vimeo90K dataset~\cite{xue2019video}.
We test the compression performance of the trained models on the Kodak image dataset containing $24$ test images.
We apply the models without any protection to the variance, while also experimenting with our proposal on top of the trained models with $5$ different values of $\epsilon$.

\begin{figure}[t]
    \centerline{\includegraphics[width=0.7\linewidth]{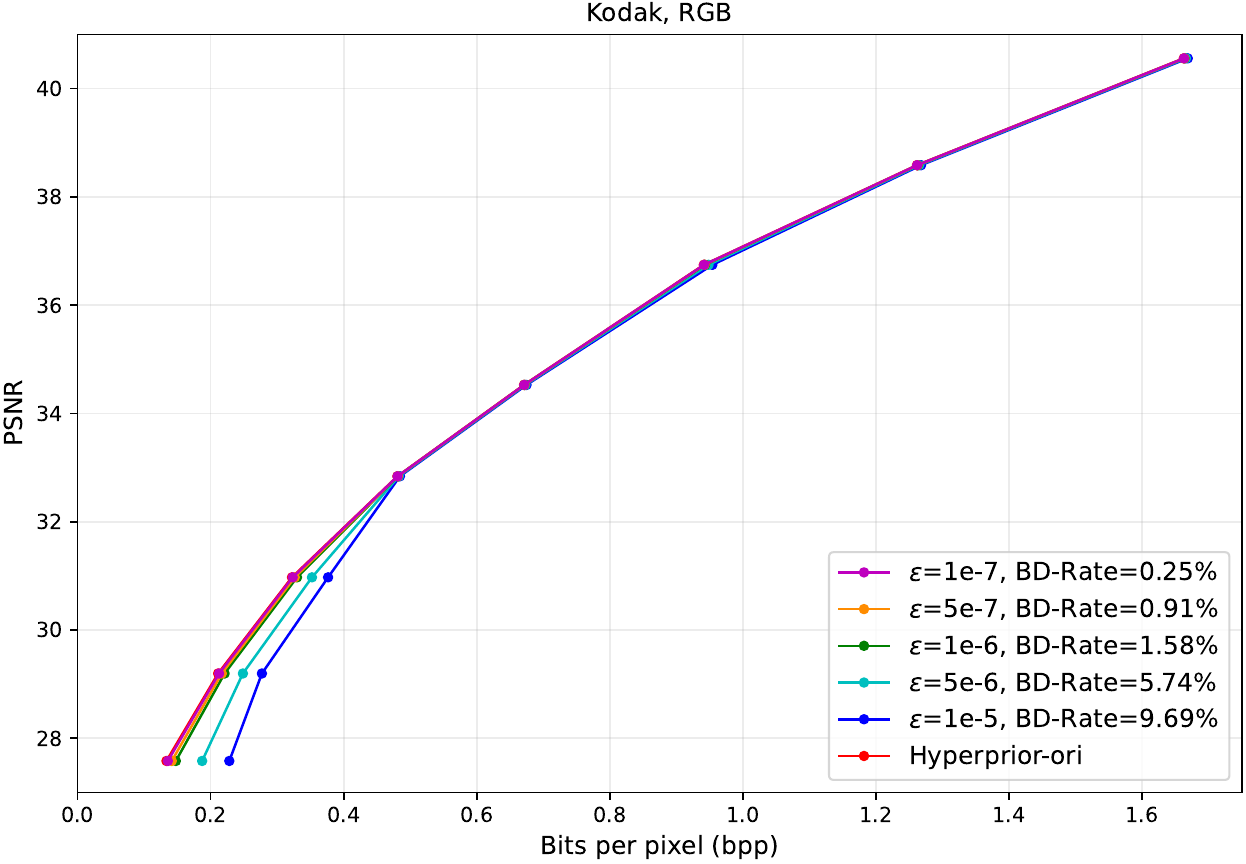}}
    \caption{R-D curves of the Kodak image dataset with different maximum tolerable errors used in our proposal.}
    \label{fig:img_comp}
    \vspace{-10pt}
\end{figure}

The R-D curves of all cases are shown in Fig.~\ref{fig:img_comp}, where the BD-Rate of our proposal compared to the original model is also labeled in the legend, indicating the percentage of the overhead when using our method.
First, by comparing the curves of our proposal and the original model (``Hyperprior-ori'' in Fig.~\ref{fig:img_comp}), we see that as the rates get higher, the safeguarding bitstream gets smaller---their differences become unnoticeable when the rates become larger than $0.5$~bpp.
Moreover, we observe that, for different curves, as $\epsilon$ decreases, the overhead clearly decreases.

\textbf{Deep octree coding}:
To inspect our proposal on bitwise deep octree coding, we select datasets ranging from dense to sparse point clouds that are recommended in the MPEG standardization activities of AI-based PCC~\cite{cfpaipcc}.
For dense point clouds, we adopt $5$ sequences from $10$- to $12$-bit: \emph{exercise}, \emph{model} ($10$-bit), \emph{basketball$\_$player}, \emph{dancer} ($11$-bit) and \emph{thaidancer} ($12$-bit).
For sparse point clouds, we adopt the \emph{KITTI} sequences $11$--$21$~\cite{Geiger2012CVPR}, then normalize and quantize them to $18$-bit point clouds.
For all the tests, we use the first $64$ frames.

Table~\ref{tab:doc} shows the overheads (in \%) of different categories, where the maximum tolerable errors $\epsilon$ and quantization step sizes $q$ are varied within each category.
First, we see that when the input point cloud gets sparser, the overhead required also becomes smaller.
That is because for sparse point clouds, very often, a well-trained network module $\mathcal{M}$ outputs estimated occupancy probabilities that are close to $0$, which does not require protection as discussed in Section~\ref{ssec:doc}.
Additionally, as the maximum tolerable error $\epsilon$ decreases, the overhead also decreases drastically; when $\epsilon=10^{-6}$, the overhead gets less than $3\%$.
Thus, it is preferred to have a module $\mathcal{M}$ with a smaller maximum achievable error $e$ so that a smaller $\epsilon$ can be picked to avoid large overhead.

\subsection{Discussions}
Given a learning-based module $\mathcal{M}$ and input data to be fed, we can theoretically tell if the mismatch is bounded over a pair of specified platforms if all implementation details are known.
In this work, we empirically verified that the errors can be bounded, and the maximum error tends to be small in practice.
For the deep octree coding tests, we run the encoder on an Nvidia Ampere 100 GPU and the decoder on a Hopper 100 GPU, while keeping track of occupancy probabilities differences between the two platforms, and found that the (empirical) maximum achievable error is around $5\times10^{-7}$.
A similar test on lossy image compression between an encoder on a Quadro 6000 GPU and a decoder on a Quadro 5000 GPU is also conducted, and a maximum achievable error of $8\times10^{-6}$ is observed.
To further reduce the error and hence the overhead from the safeguarding bitstream, we hypothesize that a shallower network is preferred.
Many techniques that promote reproducibility, like quantized network weights~\cite{gholami2022survey}, can be integrated.
We leave these aspects for future investigations.

Note that our proposal is \emph{generic}---with the same rationale, it can be applied to different learning-based compression systems.
It is also a \emph{plug-and-play} approach---there is no need to retrain or finetune the compression model.

In the end, our findings suggest that it is desired to mandate hardware/software implementations with bounded errors to facilitate reproducibility in addition to protections at the application level. Hardware/software platforms for learning-based computing are encouraged to disclose the level of possible randomness or errors for example as a new system parameter.

%------------------------------------------------------------------------
\section{Conclusion}
\label{sec:conclusion}
We provide a solution on how to ensure \emph{reproducibility} when learning-based compression systems are used across different platforms. 
A safeguarding bitstream is proposed to be generated by an encoder, while the decoder will use the safeguard bitstream to reproduce exactly the same values as in the encoder. 
Example applications of our method on lossy image compression and lossless point cloud compression demonstrate the advantages and usefulness of our proposal.


\end{document}